\documentclass[runningheads]{llncs}

\usepackage{hyperref}
\usepackage{graphicx}
\usepackage{multirow}
\usepackage{amssymb}

\begin{document}

\title{Manifold-driven Attention Maps for Weakly Supervised Segmentation}

\author{Sukesh Adiga V \and Jose Dolz \and Herve Lombaert}
\authorrunning{Adiga et al.}
\institute{ETS Montreal, Canada}

\maketitle
\begin{abstract}
Segmentation using deep learning has shown promising directions in medical imaging as it aids in the analysis and diagnosis of diseases. Nevertheless, a main drawback of deep models is that they require a large amount of pixel-level labels, which are laborious and expensive to obtain. To mitigate this problem, weakly supervised learning has emerged as an efficient alternative, which employs image-level labels, scribbles, points, or bounding boxes as supervision. Among these, image-level labels are easier to obtain. However, since this type of annotation only contains object category information, the segmentation task under this learning paradigm is a challenging problem. To address this issue, visual salient regions derived from trained classification networks are typically used. Despite their success to identify important regions on classification tasks, these saliency regions only focus on the most discriminant areas of an image, limiting their use in semantic segmentation. In this work, we propose a manifold driven attention-based network to enhance visual salient regions, thereby improving segmentation accuracy in a weakly supervised setting. Our method generates superior attention maps directly during inference without the need of extra computations. We evaluate the benefits of our approach in the task of segmentation using a public benchmark on skin lesion images. Results demonstrate that our method outperforms the state-of-the-art GradCAM by a margin of $\sim$22\% in terms of Dice score.

\end{abstract}

\section{Introduction}
Semantic segmentation is a mainstay in medical imaging, as it serves for the diagnosis and treatment of many diseases. In the last years, we have witnessed the advancements in segmentation approaches based on deep learning, mainly using Convolutional Neural Networks (CNN). This progress is partly due to the availability of large amounts of labelled training datasets \cite{dolz2018hyperdense,kamnitsas2017efficient}. Nevertheless, obtaining such large labelled data involves pixel-wise annotation of thousands of images, which is a laborious task, prone to subject-variability. This is further magnified in medical imaging since segmentation requires specific expert knowledge.

Recently, weakly supervised segmentation (WSS) has emerged as an alternative to alleviate the need for large \textit{pixel-level} labelled training datasets. These labels can come in the form of \textit{image-level} labels \cite{papandreou2015weakly}, scribbles \cite{lin2016scribblesup}, points \cite{bearman2016s}, bounding boxes \cite{rajchl2016deepcut} or direct losses \cite{kervadec2019constrained}. Among these supervisory signals, image-level labels are typically preferred, as they are easier and inexpensive to obtain \cite{bearman2016s}. This form of annotation assumes that by assigning a global label, the model will be able to find common patterns that are present in positive samples (containing the class) and do not exist on negative examples.

If learning relies entirely on image-level labels, the unique known information is the object category. In this scenario, learning discriminative features that lead to accurate pixel-level segmentation is a challenging problem, since the association between semantic categories (global) and spatial information (local) is not provided. To address this limitation, visual salient regions ---derived from complementary tasks, such as classification--- are typically integrated during training \cite{oquab2015object,pinheiro2015image}. Particularly, class activation maps (CAM) \cite{zhou2016learning} have gained popularity in identifying saliency regions based on image labels. It is achieved by associating feature maps on the last layers and weighting their activations. In practice, this boils down to replace fully connected layers in a classification network by a global average pooling (GAP) layer, which generates the class-specific feature maps, named as CAM. The main drawback of this approach is that the generated saliency maps are typically spread around the target object, only focusing on the most discriminant areas. This limits its usability as pixel-level supervision for semantic segmentation. To enhance the generated saliency regions, some alternatives based on back-propagation (GradCAM \cite{selvaraju2017grad}) or super-pixels (SP-CAM \cite{kwak2017weakly}) have been proposed. Nevertheless, these method demands additional gradients computation \cite{selvaraju2017grad} or supervision \cite{kwak2017weakly}.

The literature on weakly supervised segmentation (WSS) in medical imaging remains scarce with few alternatives to address this problem. While few methods resort to direct losses, hence requiring additional priors, such as the target size \cite{jia2017constrained,kervadec2019constrained}, other approaches rely on stronger forms of supervision, e.g., bounding boxes \cite{rajchl2016deepcut}. Tackling this from a perspective of \textit{image-level} labels typically uses visual features, which has not been much investigated \cite{feng2017discriminative,gondal2017weakly,nguyen2019novel}. For example, Nguyen et \textit{al.} \cite{nguyen2019novel} proposed CAM-based approach for the segmentation of uveal melanoma. In their method, the CAMs generated by the classification network are further refined by an active shape model and CRF \cite{krahenbuhl2011efficient}. Enhanced maps were later employed as segmentation proposals to train a segmentation CNN. More recently, CAMs derived from image-level labels were combined with attention scores to refine lesion segmentation in brain images \cite{wu2019weakly}. By doing that, they demonstrated the improvement in performance compared to the vanilla version of CAMs. However, these methods rely on a trained classification network or employ an auxiliary classification branch to generate the visual saliency regions. Thereafter, their strategy typically integrates with complex models to enhance the performance of a final segmentation. We argue that, instead, employing visual manifold networks is a new, better performing approach to discriminate identified saliency regions. Our motivation is that these networks map input images into a manifold space, where similarities between images are kept. Enforcing attention to relevant visual regions should thus lead to consistent feature representations for two different images belonging to the same class. This motivates the use of a manifold network that jointly generates robust feature representations for the manifold task and learns consistent visual attention regions of images from the same category. Also, it is not feasible to apply GradCAM directly on manifold networks \cite{chen2020adapting}, whereas our attention module in the manifold network directly produces attention maps.

\paragraph{\textbf{Our contribution.}}
We propose to derive visual attention from a manifold learning network to leverage the generated visual clues as strong proxies for semantic segmentation. Specifically, we integrate the attention module to (i) obtain visual attention directly, (ii) focus the attention on the target object for a manifold learning task, and (iii) serves as proxy labels for segmentation. As we demonstrate in our experiments, the proposed method provides better attention maps than state-of-the-art GradCAM applied on classification networks. We evaluate the proposed method with extensive experiments on a public benchmark of a skin lesion dataset, ISIC \cite{tschandl2018ham10000,codella2019skin} in the task of weakly supervised segmentation.

\begin{figure}[htb]
\centering
\includegraphics[width=0.985\linewidth]{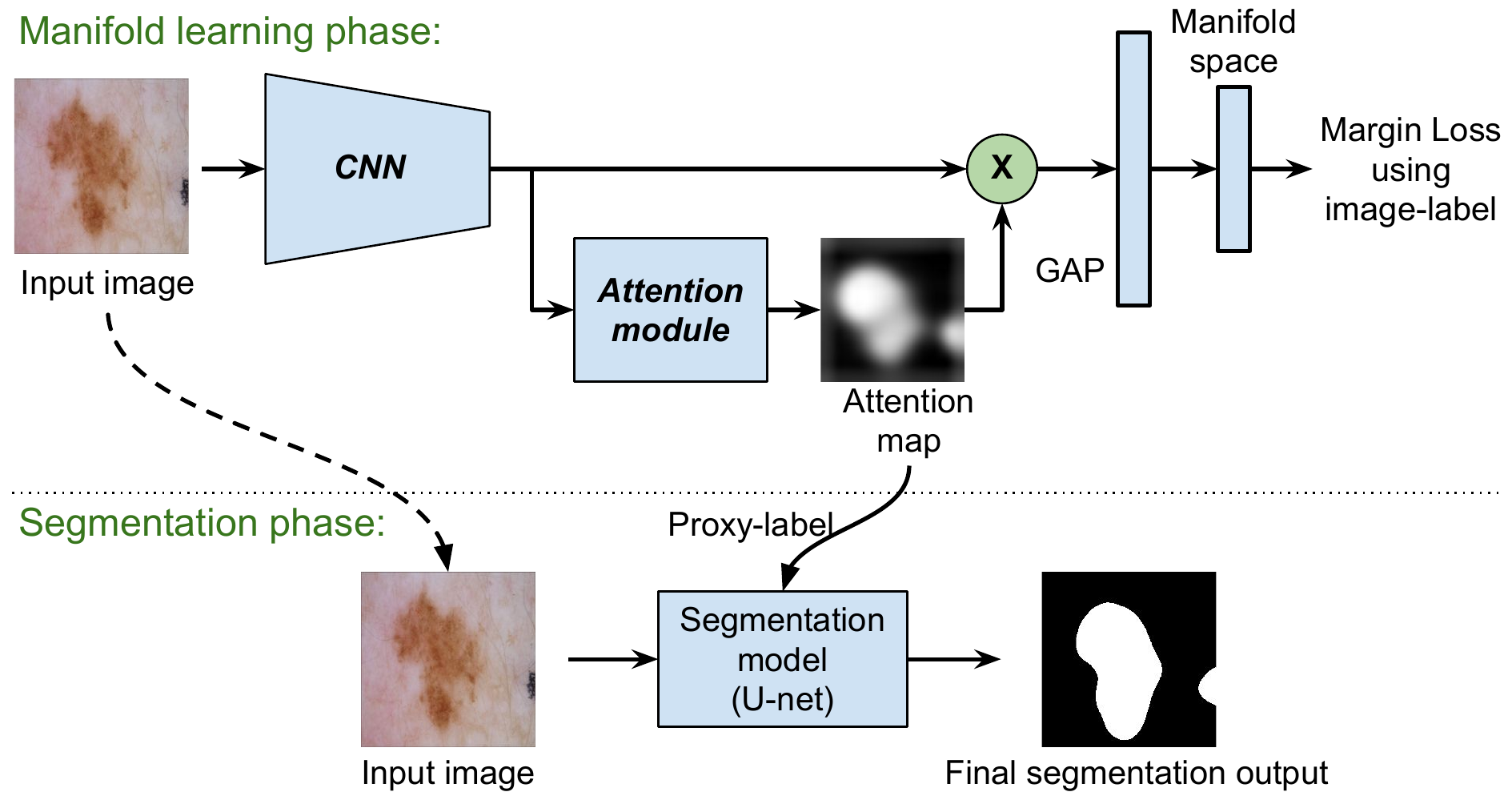}
\caption{Schematic of the proposed pipeline for weakly supervised segmentation using only image-level labels. In the manifold learning phase, the attention maps are produced while learning manifold space using image-level labels. We use these attention maps as proxy-labels in the segmentation network for pixel-level prediction.}
\label{fig:arch}
\end{figure}

\section{Method}
The pipeline of our proposed weakly supervised learning is shown in Fig~\ref{fig:arch}. The main idea is to learn attention maps from a manifold learning network trained on image-labels, which can be used as image proposal to train a segmentation CNN, mimicking full-supervision. To achieve this, we first introduce an attention module in the manifold learning pipeline, which generates an attention map for each image. The underlying manifold learning pipeline is inspired by the recent divide and conquer metric learning (DCML) method \cite{sanakoyeu2019divide}, which simplifies the learning task by dividing the original manifold space into several subspaces. The generated attention maps are then used as proxy-labels to train a segmentation network. In the following sections, we first describe our proposed attentive manifold learning formulation and weakly supervised segmentation setting.

\subsection{Attentive Manifold Learning}
Let $\{(x_i,y_i)\}^N_{i=1}$ be the training data where $x_i \in \mathbb{R}^{3 \times w \times h}$ is an image of width $w$ and height $h$, and $y_i \in \{1,2,...,C\}$ its corresponding image-level label, with $C$ as the total number of classes. Our aim is to learn the attention maps from the manifold network. To define the attention module, let $x_i$ be an input image. The feature extractor $S(\cdot)$ produces a feature map $f_i=S(x_i)$, where $f_i \in \mathbb{R}^{c \times m \times n}$. If we denote $A$ as the attention module, the attention map for a given input image $x_i$ can be defined as: 

\begin{equation}
\label{eq:attn_maps}
a_i = A(f_i); \hspace{0.3cm} a_i \in \mathbb{R}^{m \times n}.
\end{equation}

The generated attention map is multiplied with each feature map $a_i \odot f_i$, where $\odot$ is the elementwise product. This helps to focus on the target objects during the manifold learning task and facilitates the generation of an attention map directly during inference. The attentive feature maps are then combined to produce a $c-$dimensional vector by using global average pooling (GAP), which acts as a regularizer \cite{lin2013network}. The resulted features are mapped into the manifold space using a dense layer, as shown in Fig~\ref{fig:arch}.

To learn the manifold space, we employ a metric learning approach, i.e. $F_\theta(x_i): \mathbb{R}^{3 \times w \times h} \rightarrow \mathbb{R}^d$, where $d$ is the dimension of manifold space. Metric learning maps the semantically similar images in the input space $\mathbb{R}^m$ (i.e., same class) onto metrically close points in the learned manifold $\mathbb{R}^d$. Similarly, semantically dissimilar images in $\mathbb{R}^m$ should be mapped metrically far in $\mathbb{R}^d$. The parameters $\theta$ are typically learned using a distance metric. In this work, we use, without loss of generality, a Margin loss \cite{wu2017sampling} as a distance metric to learn the parameters \footnote{Note that any other distance metric can be used as the loss function for this task.}, defined as:

\begin{equation}
\label{eq:margin}
l_{margin}(x_i, x_j) = [\alpha + \mu_{ij} (d(F_\theta(x_i), F_\theta(x_j)) - \beta)]_+, 
\end{equation}
where $d(F_\theta(x_i), F_\theta(x_j))$ is Euclidean norm between a pair of images $x_i$ and $x_j$ in the manifold space $\mathbb{R}^d$. The parameters $\alpha$ and $\beta$ represent the separation margin and the boundary between the similar and dissimilar pairs, respectively. The parameter $\mu_{ij} \in \{-1, 1\}$ indicates whether the images in the pair are similar ($\mu_{ij}=1$) or different ($\mu_{ij}=-1$). 

Several metric learning methods have been explored for learning the manifold space. We follow the recent state-of-the-art metric learning method in \cite{sanakoyeu2019divide}. This method is motivated by the idea of \textit{divide and conquer} approach, which divides a complex problem into several easier subproblems. Particularly, this method splits the manifold space $\mathbb{R}^d$ and the data into multiple groups and learns each subspace with independent learners. We adopt this method for medical imaging and integrate the attention module for better learning the manifold space and thereby enhancing the derived attention maps.

\subsection{Weakly Supervised Segmentation}
The attention maps obtained from the manifold network using \textit{image-level} labels can serve as \textit{pixel-level} labels. To further refine the attention maps, a segmentation network is trained on the fake \textit{image-level} labels. Specifically, we use the input image $x_i$ and its corresponding generated attention map $a_i$ as a training pair. To differentiate foreground pixels from the background pixels, we threshold the attention maps with $T$ (i.e., pixels in $a_i$ greater than T are set to 1, 0 otherwise) before training the segmentation network. We choose the popular segmentation network U-net \cite{ronneberger2015u} for our experiments. The network is trained with cross-entropy as a loss function, which is computed over a pixel-wise soft-max activation on the final feature maps, defined as
\begin{equation}
\mathcal{L}_{CE}(x, a) = - \sum_{j=1}^{P} a^j log(F_{\theta}(x)^j))
\end{equation}
where $F_{\theta}$ is a segmentation network parameterized by $\theta$, and $P$, the number of categories.

\section{Experiments}
The performance of the proposed attention-based approach for weakly supervised segmentation is compared with GradCAM \cite{selvaraju2017grad}, as it has been applied for medical image segmentation. We generate the GradCAMs for two standard classification networks based on ResNet50 and ResNet101. Since we employ the divide and conquer approach (DCML) \cite{sanakoyeu2019divide} for the underlying manifold learning pipeline, we compare with the standard metric learning (ML) method, which is trained using margin loss \cite{wu2017sampling}. We also include the results of full-supervision using U-net \cite{ronneberger2015u}, which serves as an upper bound. For a meaningful evaluation, the model architecture and other parameters are fixed across the different methods, as described in Sec.~\ref{sec:impn}. In the following sections, the dataset composition used for training and evaluation, as well as the implementation details of our pipeline are detailed. Then, we present the quantitative and qualitative results of the proposed approach comparing with the baseline methods for weakly supervised segmentation.

\subsection{Datasets}
The proposed method is evaluated on the skin lesion dataset from the ISIC 2018 Challenge \footnote{https://challenge2018.isic-archive.com/} \cite{tschandl2018ham10000,codella2019skin}. The dataset consists of two independent sets. The first dataset contains 10,015 images with seven different categories for classification. The second dataset focuses on the segmentation task and is composed of 2,594 images and their corresponding pixel-level masks. We use the classification data to generate attention maps by learning the manifold space. To this end, the dataset is split into independent 8,015 images for training and 2,000 images for testing. For the segmentation task, we leverage the attention maps generated from the classification set (i.e. 10,015 images), which are employed as proxy-labels to train the segmentation network. The segmentation dataset is randomly split into three sets: training (1,042), validation (520), and testing (1,038). We employ the validation and testing splits to evaluate all the methods. In contrast, the training set is used to train the upper-bound model, i.e., full-supervised.

\subsection{Implementation details}
\label{sec:impn}
We follow the work in \cite{sanakoyeu2019divide} as the backbone architecture for learning the manifold space, which is based on ResNet-50 \cite{he2016deep}. From this network, we use only three residual blocks to avoid a low resolution on the generated attention modules. The attention module consists of three convolution layers with $3 \times 3$ kernel and filters size of \{128, 32, 1\}. This module integrates a ReLU activation between the convolutional layers and a sigmoid activation in the final layer to produce an activation map. The manifold dimension size is fixed to $d$ = 128 and an input image size of $224 \times 224$ used for all our experiments. All models are trained using the Adam optimizer \cite{kingma2014adam} with batch sizes of $B$ = 32. The margin loss parameters are set to $\alpha$ = 0.2 and $\beta$ = 1.2, as in \cite{wu2017sampling}. In each mini-batch, 8 images per class are sampled to ensure a class-balanced scenario and experiments are trained for 300 epochs. The last 50 epochs are fine-tuned in the full embedding. For the segmentation network, we use U-net \cite{ronneberger2015u} architecture with an initial kernel size of 32. It is also trained with Adam optimizer with batch sizes of 16 for the binary segmentation ($P=2$). The threshold parameter is set to $T=0.5$ for all the experiments.

\subsection{Evaluation of Segmentation using Dice Score}
\begin{table}[ht!]
\centering
\addtolength{\tabcolsep}{5.8pt}
\begin{tabular}{l | c | c | c | c }
\hline \hline
 set                & \multicolumn{2}{c|}{validation}     & \multicolumn{2}{c}{test} \\ [0.5ex]
 \hline
 Method             & init maps & U-net & init maps & U-net \\ [0.5ex]
 \hline \hline
 GradCAM $^\ast$     & 34.80    & 41.12   & 34.00     & 40.65  \\
 GradCAM $^\dagger$     & 34.16    & 39.03   & 33.68     & 39.53  \\
 \hline
 ML + Attention  & 56.60	& 58.10  & 56.96 & 59.16 \\
 DCML + Attention (ours)    & \textbf{60.79} & \textbf{63.83} & \textbf{62.06} & \textbf{66.12}  \\
 \hline
 Full-supervision (upperbound) & - & 85.90 & - & 86.15  \\ [0.5ex]
 \hline \hline
\end{tabular}
\caption{Quantitative comparison using Dice score (in \%) on validation and test sets. Our method yields the best results, in bold for the weakly supervised setting. $^\ast$ and $^\dagger$ are obtained by GradCAM on classification networks using ResNet50 and ResNet101, respectively.}
\label{table:dicescore}
\end{table}

We employ the Dice score to evaluate the segmentation performance of the proposed method along with baseline approaches. Table~\ref{table:dicescore} reports these results for the validation and testing datasets. In this table, \textit{init maps} are used to denote the raw visual salient regions from either GradCAM or the proposed method. On the other hand, U-net refers to the performance of the segmentation network trained on the \textit{init maps}. First, we can observe that segmentation results obtained with the initial GradCAM are considerably low. Particularly, on both validation and testing sets, both variants of GradCAM (ResNet50 and ResNet101) achieve a Dice score of around $\sim$34\%. If these raw maps are used as proxy image-labels to train a segmentation network, results are improved by $\sim$6\%. However, even in this case, the performance is still insufficient, with a maximum Dice score of 41.12\%. The attention maps produced in standard metric learning represent better segmentation compared to the GradCAM variants, as it achieves a Dice score of 56.6\% and 56.96\% on validation and test sets, respectively. The performance of this model is further improved by $\sim$2\% if they are used to train a segmentation network. Last, we can observe that our method based on DCML achieves the best Dice score of 60.79\% and 62.06\% for raw attention maps. Furthermore, the Dice score is further improved by $\sim$3\% and $\sim$4\% on validation and test set, when the attention maps are used as proxy-labels. Compared to GradCAM, our visual manifold driven methods shows superior performance with an improvement of $\sim$22\% Dice score due to the similarity-based metric learning. In addition, compared to the standard metric learning, our method (DCML) brings a gain of performance between 4-7\% due to the subspace learning. This suggests that the proposed model generates more reliable segmentations that can be further employed to train fully-supervised networks.

\begin{figure}[ht!]
\centering
\includegraphics[width=0.985\linewidth]{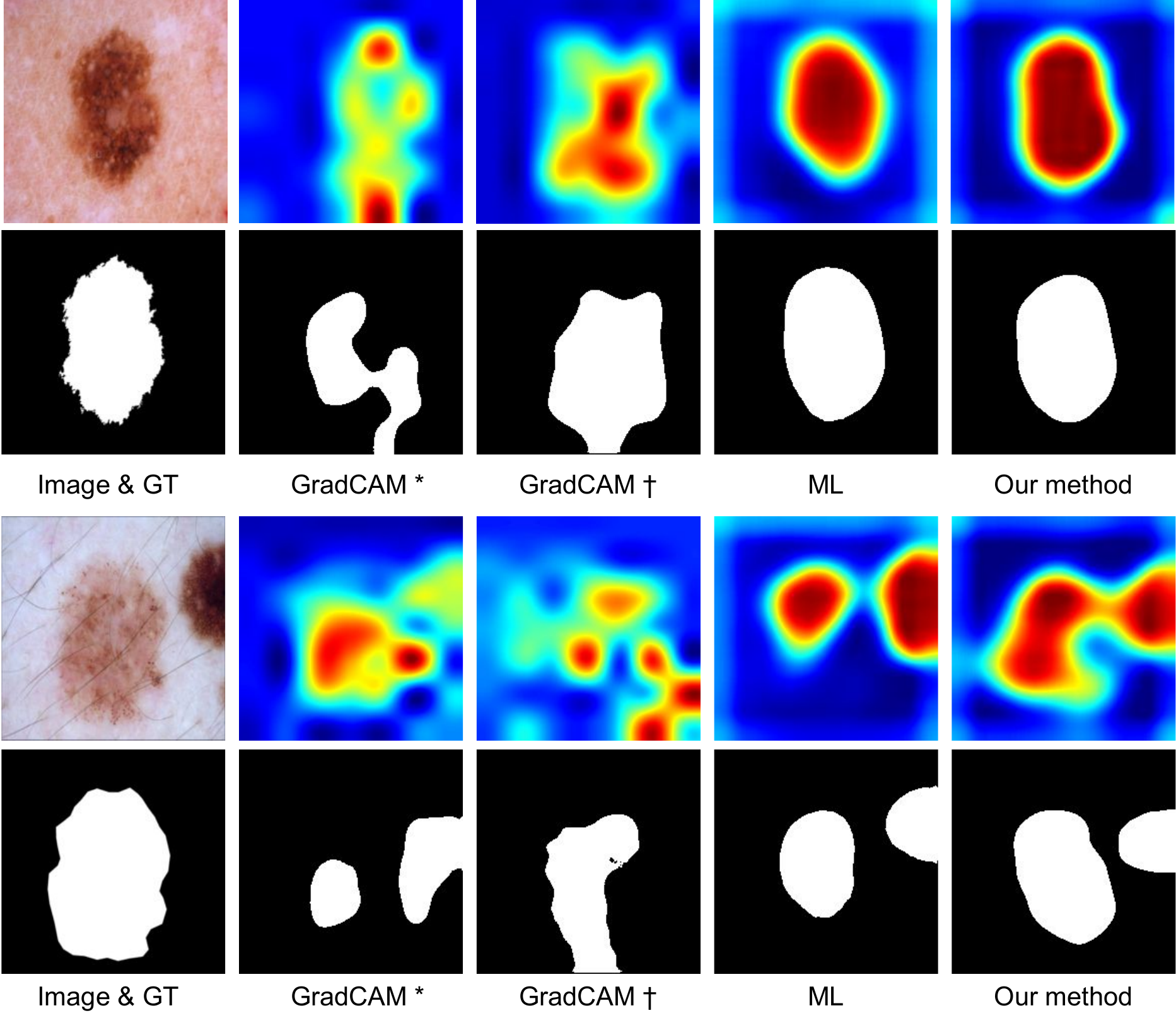}
\caption{Saliency map obtained by different method (row 1 and 3) and segmentation results obtained in weakly supervised setting (row 2 and 4). $^\ast$ and $^\dagger$ are obtained by GradCAM on classification networks using ResNet50 and ResNet101, respectively.}
\label{fig:seg}
\end{figure}

\subsection{Qualitative Performance Evaluation}
Visual results of the different methods are shown in Fig. ~\ref{fig:seg}. The saliency maps (row 1 and 3) produced by GradCAM spread over the entire image, highlighting discriminative regions of the lesion but failing to capture the whole context. In contrast, attention maps derived from metric learning better capture the attentive region, mostly covering the lesion region. 
This shows the potential of attention maps generated by the manifold learning over GradCAM on classification networks. Additionally, compared to standard metric learning, our method captures finer details, which may be due to the multiple-subspace learning, which eases the task. The segmentation results obtained by training a segmentation network on the initial salient regions (row 1 and 3) are depicted in row 3 and 4. These images demonstrate the feasibility of our method to weakly generate pixel-level labels that can be used to train segmentation networks.

\section{Conclusion}
We presented a novel manifold-driven attention-based pipeline for weakly supervised segmentation using image-level labels. Our method directly produces an attention map, which serves as proxy labels for segmentation. The segmentation results outperform the state-of-the-art GradCAM methods by a margin of $\sim$22\% Dice score, for an application on skin lesion images. Qualitative results demonstrate that both attention map and segmentation by our method, focusing on the target lesion, showing the effectiveness and robustness of our method. Our proposed pipeline can be easily fit in any complex weakly supervised setting, which can be explored in future work. 

\section*{Acknowledgments}
This research work was partly funded by the Natural Sciences and Engineering Research Council of Canada (NSERC), Fonds de Recherche du Quebec (FQRNT), and NVIDIA Corporation with a donation of a GPU.

\bibliographystyle{splncs04}
\bibliography{ml_wss.bbl}
\end{document}